\documentclass[10pt,twocolumn,letterpaper]{article}

\usepackage[pagenumbers]{wacv} 
\usepackage{bbold}          

\graphicspath{ {./images/} }

\definecolor{wacvblue}{rgb}{0.21,0.49,0.74}
\usepackage[pagebackref,breaklinks,colorlinks,allcolors=wacvblue]{hyperref}

\def\wacvPaperID{####} 
\def\confName{WACV}
\def\confYear{2026}

\title{SurfDist: Interpretable Three-Dimensional Instance Segmentation Using Curved Surface Patches}

\author{Jackson Borchardt\\
Weill Institute of Neurosciences \\
University of California, San Francisco\\
{\tt\small borc -at- berkeley.edu}
\and
Saul Kato\\
Weill Institute of Neurosciences \\
University of California, San Francisco\\
{\tt\small saul.kato -at- ucsf.edu}
}

\begin{document}
\maketitle
\begin{abstract}

We present SurfDist, a convolutional neural network architecture for three-dimensional volumetric instance segmentation. SurfDist is a modification of the popular model architecture StarDist-3D which enables learning instance boundaries as closed piecewise compositions of smooth parametric surfaces. This parameterization breaks StarDist-3D's coupling of instance dimension and instance voxel resolution, and it produces predictions which may be upsampled to arbitrarily high resolutions without introduction of voxelization artifacts. For datasets with blob-shaped instances, common in biomedical imaging, SurfDist can achieve higher segmentation accuracy than StarDist-3D with more compact instance parameterizations.

\end{abstract}

\section{Introduction}
\label{sec:intro}

In image segmentation, machine learning approaches have achieved major advances over classical computer vision techniques. However, these approaches pose mechanistic interpretability challenges that can make it difficult both to reason about a model's ability to generalize beyond its training dataset and to value synthetic data generated by a trained model. Additionally, instance segmentation of 3D volumetric data remains undeveloped compared to instance segmentation of 2D images, because of additional computational expense associated with 3D machine learning models as well as a lack of publicly available high-quality 3D benchmark datasets.

The StarDist family of models has become popular for biomedical image segmentation tasks where instances are blob-like objects. These models are constructed by augmenting "backbone" instance segmentation networks, such as a 3D U-Net \cite{cicek20163dunetlearningdense}, with additional output layers which enforce instance embeddings into an explicitly parameterized geometric space. This construction both aids interpretability and improves segmentation accuracy relative to backbone networks alone.

\begin{figure}
\centering
\begin{subfigure}{0.275\textwidth}
    \includegraphics[width=\textwidth]{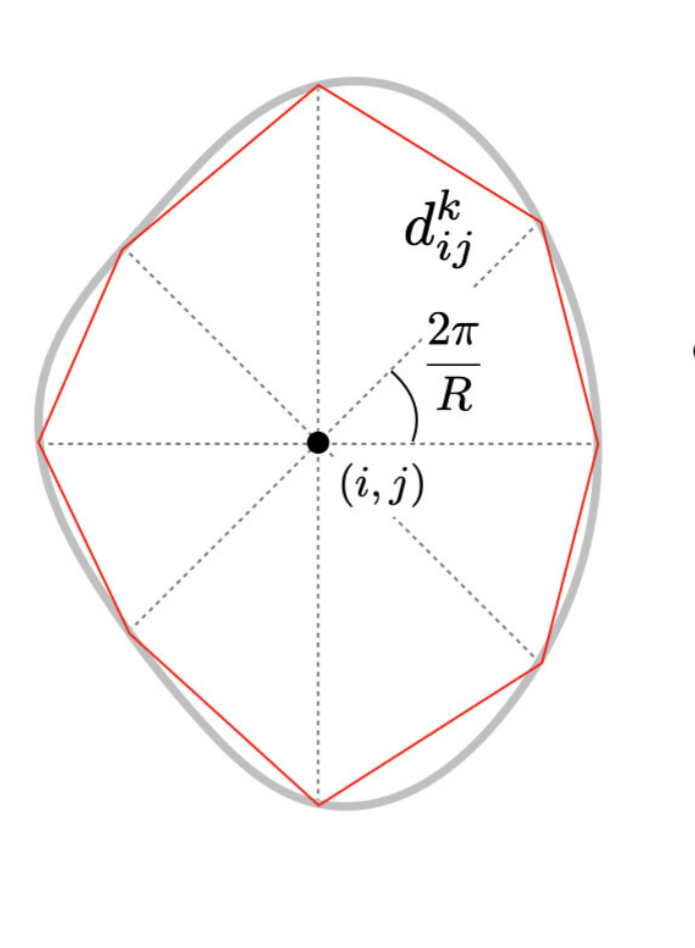}
    \subcaption{An example 8-ray StarDist instance. StarDist models instances as sets of radial distances along equiangular directions (black lines) from object centers to object boundary intersections. These intersection points define a polygon (red) which approximates the object boundary (gray).}
    \label{fig:star}
\end{subfigure}
\begin{subfigure}{0.275\textwidth}
    \includegraphics[width=\textwidth]{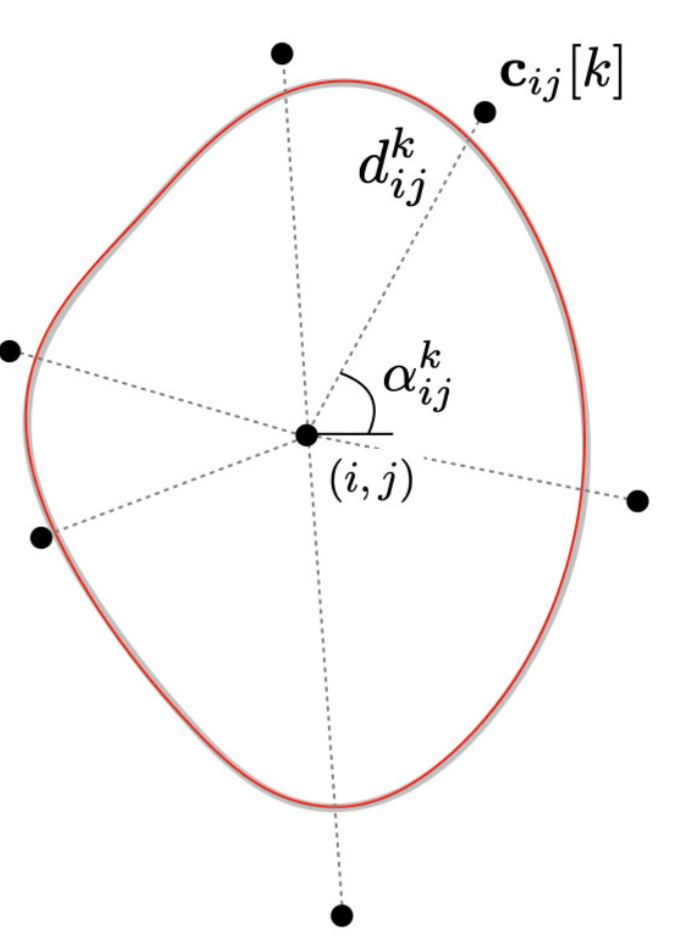}
    \subcaption{A 6-control point SplineDist instance corresponding to \cref{fig:star}. SplineDist models instances as sets of spline basis points (black) which together define smooth closed object contours (red).}
    \label{fig:spline}
\end{subfigure}
\caption{Representative StarDist and SplineDist instance models from \cite{mandal2021splinedist}.}
\label{fig:2d_meshes}
\end{figure}

In the original StarDist model \cite{schmidt2018cell}, instances within 2D images are modeled as star-convex polygons (\cref{fig:star}). This parameterization often works well in practice, but because it predicts piecewise linear approximations of instance boundaries, its ability to model local curvature degrades as instance sizes (in pixels) increase. The SplineDist model \cite{mandal2021splinedist} solves this problem by modifying StarDist to model instances as closed spline curves (\cref{fig:spline}) rather than as polygons. This approach decouples model complexity from instance spatial resolution, and empirically, it reduces, relative to StarDist, the number of network parameters required to achieve a given level of segmentation accuracy.

StarDist-3D \cite{weigert2020star} extends the 2D StarDist approach to the 3D setting by modeling instances as star-convex polyhedra. Like the 2D StarDist model, it works well for many current datasets but couples local curvature modeling to instance spatial extent. Toward breaking this coupling in the 3D setting, we present SurfDist, a modification of StarDist-3D which models instances as contiguous, closed meshes of bicubic triangular surface patches. We find that SurfDist achieves segmentation accuracy comparable to or better than that of StarDist-3D on real biological data while fitting simpler instance representations.
\section{Technical details}

\subsection{Mesh parameterization}

\begin{figure}
    \centering
    \includegraphics[width=0.35\textwidth]{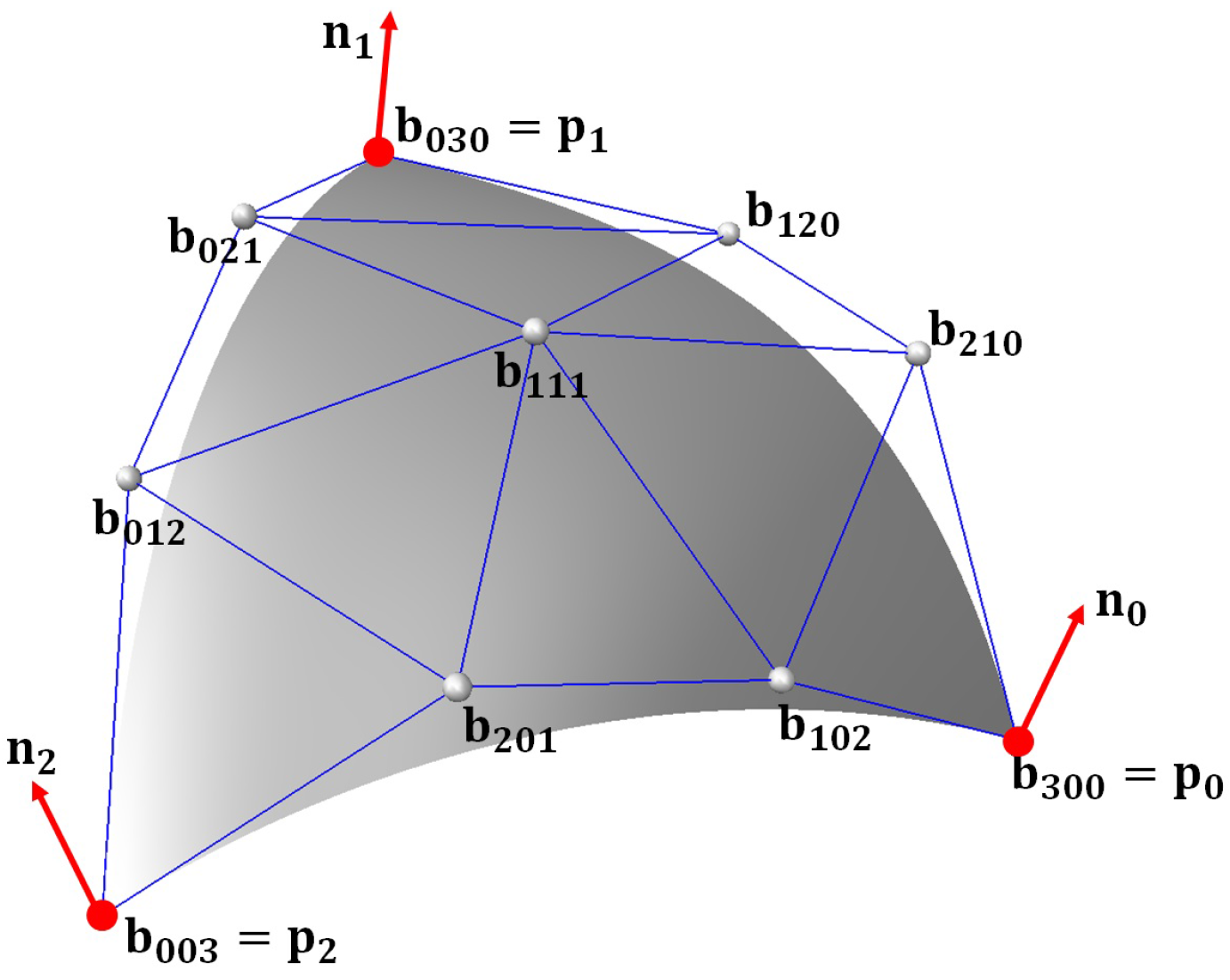}
    \caption{A generic bicubic Bézier triangle from \cite{lee2016bezier}. Control points are annotated with gray spheres, vertex control points are annotated with red spheres, vertex normals are annotated with red arrows, the control mesh is shown in blue, and the smooth triangular surface parameterized by the control points is rendered in gray.}
    \label{fig:bezier}
\end{figure}

SurfDist models instances as meshes of curved triangular surface patches. Each patch is formulated as a bicubic Bézier triangle parameterized by 10 control points (\cref{fig:bezier}). Of these 10 control points, 3 are at the triangle's vertices (and are guaranteed to be on the resulting surface), 6 are associated with (but generally not on) the triangle's edges, and 1 is associated with (but generally not on) the triangle's interior. Points on the patch surface are given by the bicubic interpolation 

\begin{equation}
    p(u,v,w) = \sum_{i=0}^{3} \sum_{j=0}^{3-i} \frac{3!}{i!j!k!} u^i v^j w^k b_{ijk}^{3}
  \label{eq:triangle}
\end{equation}

across a barycentric triangular input domain $\{u,v,w \mid 0 \leq u,v,w \leq 1, u+v+w = 1\}$ where $\mathbf{b}$ is the set of control points as indexed in \cref{fig:bezier} and $k = 3-i-j$ (\cite{alma991085856964006532}).

Since SurfDist meshes comprise multiple triangular patches, and since adjacent patches share vertices and edges, all control points except those associated with triangle interiors are shared by multiple triangles in a given mesh. Thus, the total number of control points for a given mesh is given by $V + 2E + T$ where $V$ is the number of vertices, $E$ is the number of edges, and $T$ is the number of triangles in the mesh. As in StarDist-3D, $V$ is the only value of these three which is a specifiable hyperparameter. $E$ and $T$ are determined by the triangular mesh given by taking the convex hull of a mapping of each vertex in $V$ to a position on the unit sphere using either the Fibonacci lattice \cite{gonzalez2010measurement} or an equidistant spacing scheme.

\begin{figure}
\centering
\begin{subfigure}{0.45 \textwidth}
    \includegraphics[width=\textwidth]{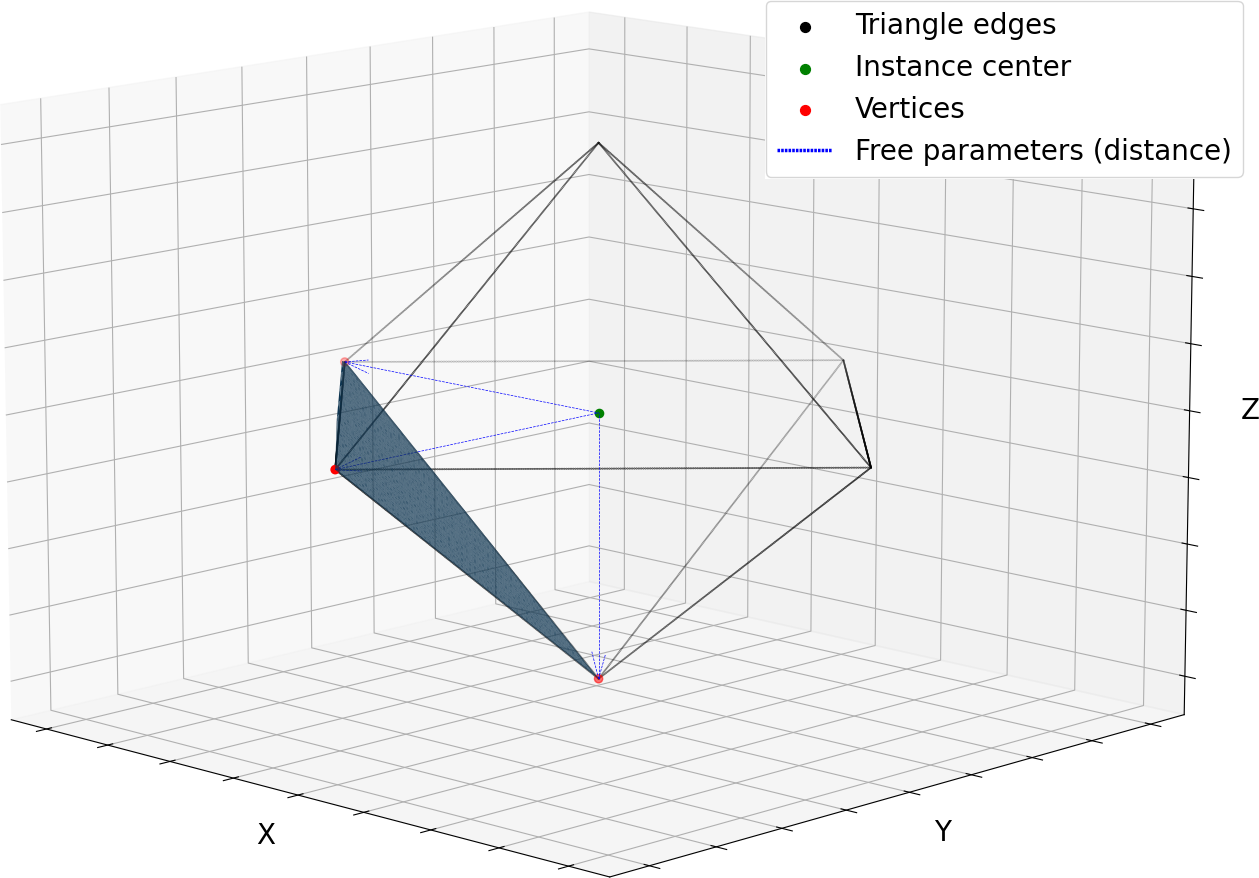}
    \caption{An example 6-ray StarDist-3D instance. Vertices of a single arbitrarily chosen polyhedral facet are annotated in red, and the surface for that facet is rendered in blue. The radial distances from the center of the instance to the six vertices of the polyhedron are the only free parameters of this representation.}
    \label{fig:star3dmesh}
\end{subfigure}
\begin{subfigure}{0.45 \textwidth}
    \includegraphics[width=\textwidth]{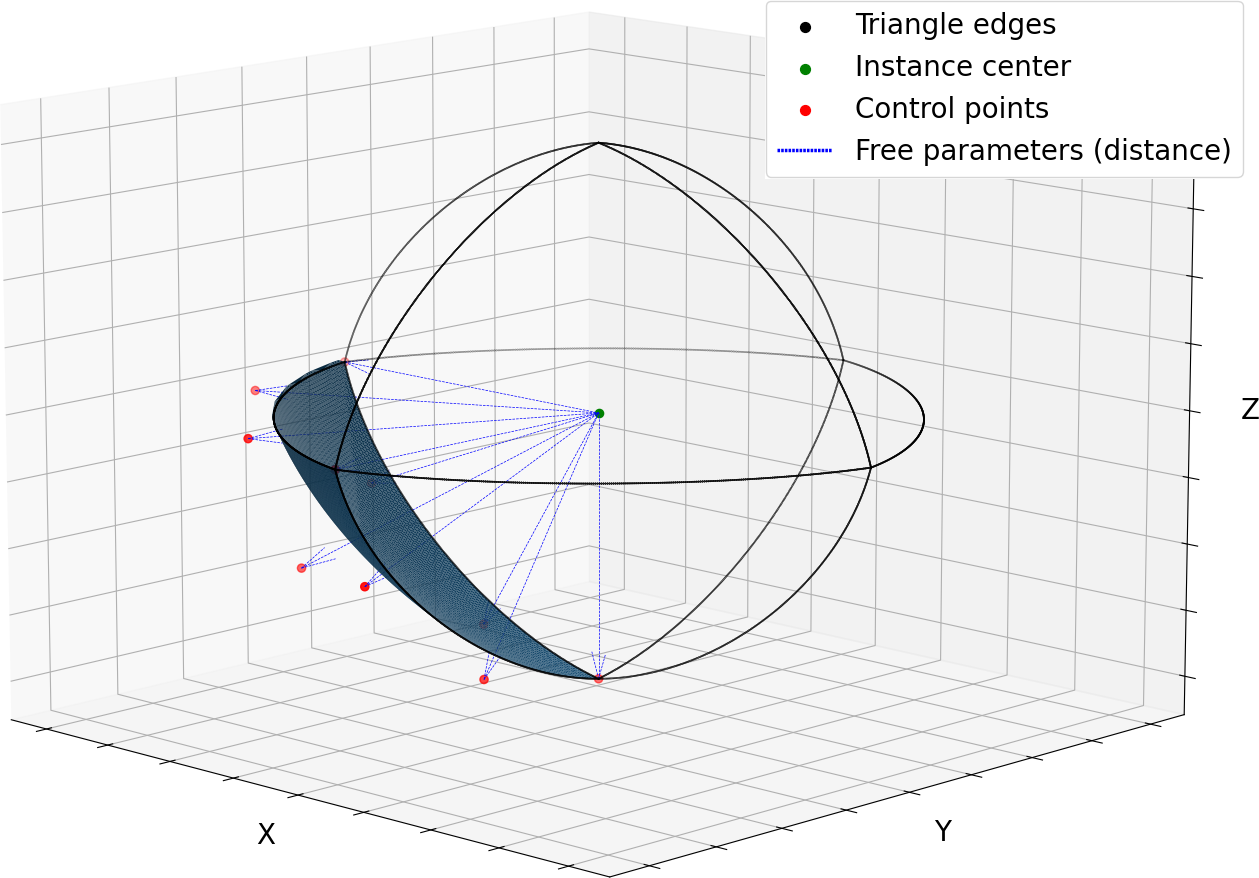}
    \caption{A 6-ray SurfDist instance corresponding to \cref{fig:star3dmesh}. Control points of a single arbitrarily chosen Bézier triangular patch are annotated in red, and the patch surface is rendered in blue. The radial distances from the center of the instance to the 38 control points (6 vertex control points, 24 edge control points, and 8 interior control points) of the triangular mesh are the free parameters of this representation.}
    \label{fig:surfmesh}
\end{subfigure}
\caption{A StarDist-3D versus a SurfDist model instance. The models use the same number of mesh vertices and triangular faces, but the SurfDist model is better suited to reproducing smooth instance surfaces.}
\label{fig:3d_meshes}
\end{figure}

Once a triangular mesh has been calculated, the control points for each triangle are assumed to lie along the rays drawn from the unit sphere's center toward the triangle's vertices (for vertex control points), toward its center (for interior control points), or toward points that divide its edges into thirds (for edge control points). Instance predictions then simply comprise radial distance estimates for each control point, so the number of free parameters for a given instance is equal to the number of control points in its mesh (\cref{fig:surfmesh}). This construction results in a topologically closed mesh which is smooth across triangular surfaces and continuous across the edges of constituent triangles. Importantly, mesh surfaces and edges may be iteratively subdivided (\cite{kato2002curved}). This property enables direct use of StarDist-3D's inference implementation (by using subdivision to approximate curved meshes with arbitrarily high-detailed polyhedra), and it allows for sampling of surface points at arbitrarily fine resolutions during both model training and inference.

\subsection{Loss}

Loss within StarDist family models is formulated as a sum of object ($L_{\mathit{obj}}$) and distance ($L_{\mathit{dist}}$) terms. The object term is a probability, and the distance term has absolute magnitude. These terms are defined at the voxel level: for a given voxel, the object term tracks foreground identity confidence, and the distance term tracks predicted instance fit. Loss for a given input image is calculated by taking the mean absolute error (MAE) of loss across all of its voxels. Formally, per-voxel loss is defined as

\begin{equation}
  L(p, \hat{p}, \mathbf{d}, \mathbf{\hat{d}}) =  L_{\mathit{obj}}(p, \hat{p}) + \lambda_d L_{\mathit{dist}}(p, \mathbf{d}, \mathbf{\hat{d}})
  \label{eq:loss}
\end{equation}

where $p$ and $\hat{p}$ are actual and predicted foreground probability measures calculated as distance to the nearest exterior (either background or interior of another instance) voxel normalized by the maximum such distance for the voxel's instance, $L_{\mathit{obj}}$ is standard binary cross-entropy loss, $\lambda_d$ is a specifiable regularization weight, $\mathbf{d}$ and $\mathbf{\hat{d}}$ are vectors of predicted and actual radial distances to instance exterior along $n$ radial directions, and

\begin{equation}
  L_{\mathit{dist}}(p, \mathbf{d}, \mathbf{\hat{d}}) =  p \cdot \mathbb{1}_{p>0} \cdot \frac{1}{n}\sum\nolimits_k |  d_k -\hat{d}_k| + L_{\mathit{bg}}(p, \mathbf{\hat{d}})
  \label{eq:loss_dist}
\end{equation}

and

\begin{equation}
  L_{\mathit{bg}}(p, \mathbf{\hat{d}}) = \lambda_{\mathit{reg}} \cdot \mathbb{1}_{p=0} \cdot  \frac{1}{n}\sum\nolimits_k   |\hat{d}_k|
  \label{eq:loss_background}
\end{equation}

where $\lambda_{\mathit{reg}}$ is another specifiable regularization weight.

These loss equations are identical to the ones used in StarDist-3D (\cite{weigert2020star}), but there are two important differences between StarDist-3D's and SurfDist's loss formulations related to the set of radial directions $\mathbf{K}$. First, in StarDist-3D, $\mathbf{K}$ is determined programatically at model instantiation time using a spherical Fibonacci lattice (\cite{gonzalez2010measurement}) and observed instance anisotropy across the training dataset, and it is then held constant across all voxels for all training and inference steps. In SurfDist, $\mathbf{K}$ is determined on a per-voxel and per-prediction basis as 

\begin{equation}
    \mathbf{K} = \{\frac{\mathbf{s_1} - \mathbf{c}}{|\mathbf{s_1} - \mathbf{c}|},...,\frac{\mathbf{s_n} - \mathbf{c}}{|\mathbf{s_n} - \mathbf{c}|}\}
    \label{eq:radial_directions}
\end{equation}

where $\mathbf{S}$ is a set of surface points sampled from the smooth triangular faces of the mesh predicted by the model for a given voxel and $\mathbf{c}$ is the coordinate vector of that voxel.

Second, in StarDist-3D, the number of rays $n$ used to calculate the $L_{\mathit{dist}}$ term is defined by the number of parameters in the output layer of the model. In SurfDist, $n$ is determined by the surface point sampling paradigm (which is a hyperparameter), and increasing or decreasing the value of $n$ does not change the parameter count of the trained model (although it will affect memory requirements for model training steps). This means that even low parameter-count SurfDist models may be trained to predict entire instance surfaces. In SurfDist, the set of surface points whose barycentric coordinates are generated using two iterations of standard triangle subdivision, where one triangle is divided into four. We find that this sampling paradigm provides a reasonable balance between geometric detail and resource requirements during model training.
\section{Experimental results}

We present results for one synthetic toy dataset, \textit{Sphere}, and three publicly available volumetric microscopy datasets, \textit{Worm} \cite{long20093d}, \textit{Parhyale} \cite{alwes_2023_8252039}, and \textit{NucMM} \cite{lin2021nucmm}.

\subsection{Reconstruction of Synthetic Spheres}

\begin{figure}
  \centering
  \includegraphics[width=0.475\textwidth]{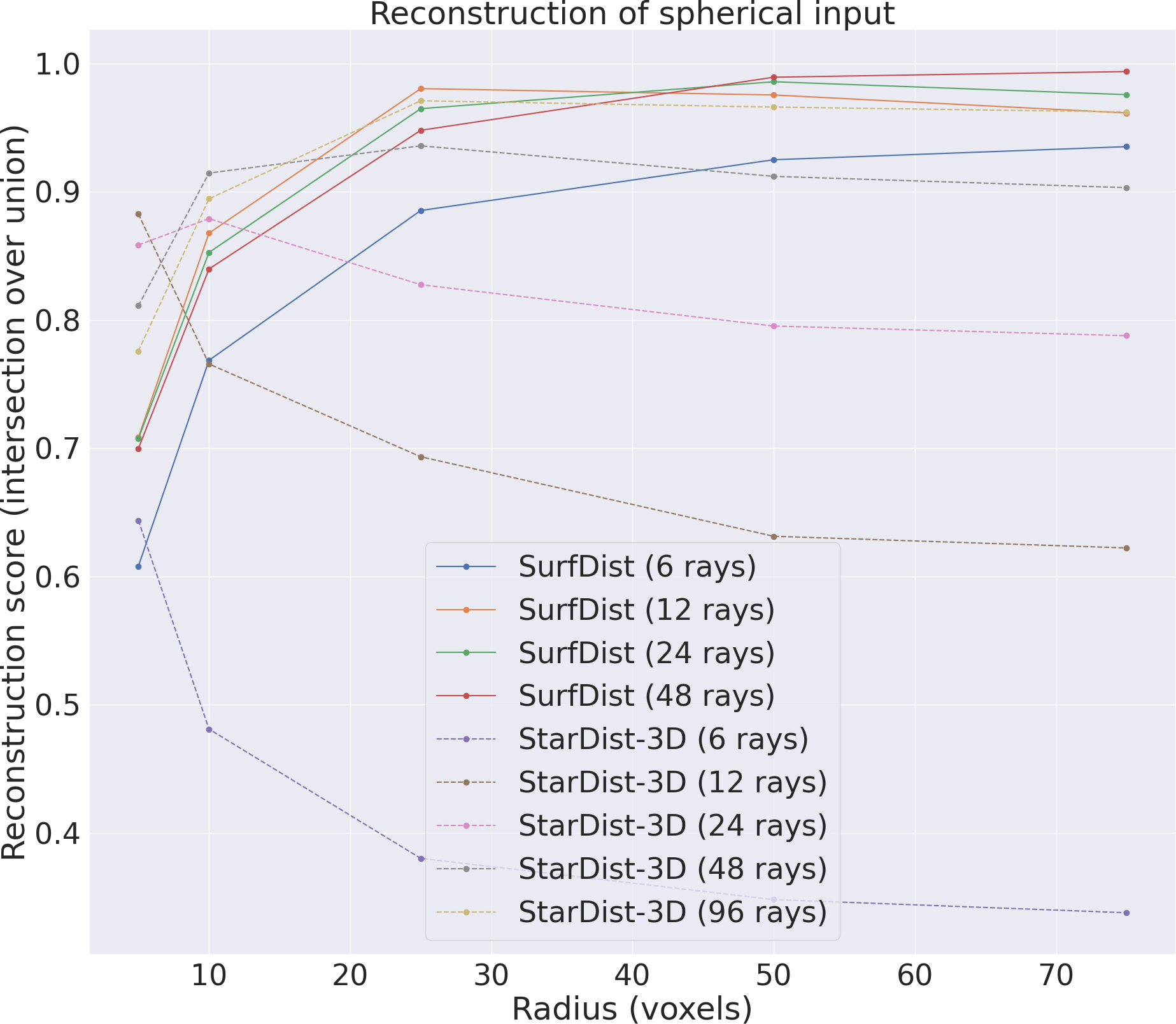}
  \caption{
  StarDist-3D vs. SurfDist reconstruction error for a voxelized sphere. With fewer parameters, SurfDist better models large spheres than StarDist-3D.
  }
\label{fig:sphere_reconstruction}
\end{figure}

\textit{Sphere} was constructed to trivially validate SurfDist's modeling approach by illustrating its ability to model objects with smooth surfaces at higher fidelity and with fewer parameters than StarDist-3D. Each of its images is a single centered spherical mask voxel set (generated by \cite{scikit-image}'s \texttt{morphology.ball} function) with spherical radial lengths varying across images. For each image, we compared reconstruction accuracy between SurfDist and StarDist-3D instance parameterizations across a range of instance ray counts. For StarDist-3D reconstructions, radial distances from image centers were measured directly. For SurfDist reconstructions, vertex control point distances were set using spherical radial lengths, and face and edge control point distances (which were assumed to be the same across all face and across all edge control points for a given image) were optimized using \cite{2020SciPy-NMeth}'s \texttt{optimize.least\_squares}. Our results (\cref{fig:sphere_reconstruction}) show that as instance size increases, SurfDist's parameterization is much better at reconstructing spherical input than StarDist-3D's. Generally, this is true even when SurfDist is allowed many fewer parameters than StarDist-3D.

\subsection{Segmentation of Worm Volumetric Microscopy Dataset}

\begin{figure*}
    \centering
    \begin{subfigure}{0.475\textwidth}
        \includegraphics[width=\textwidth]{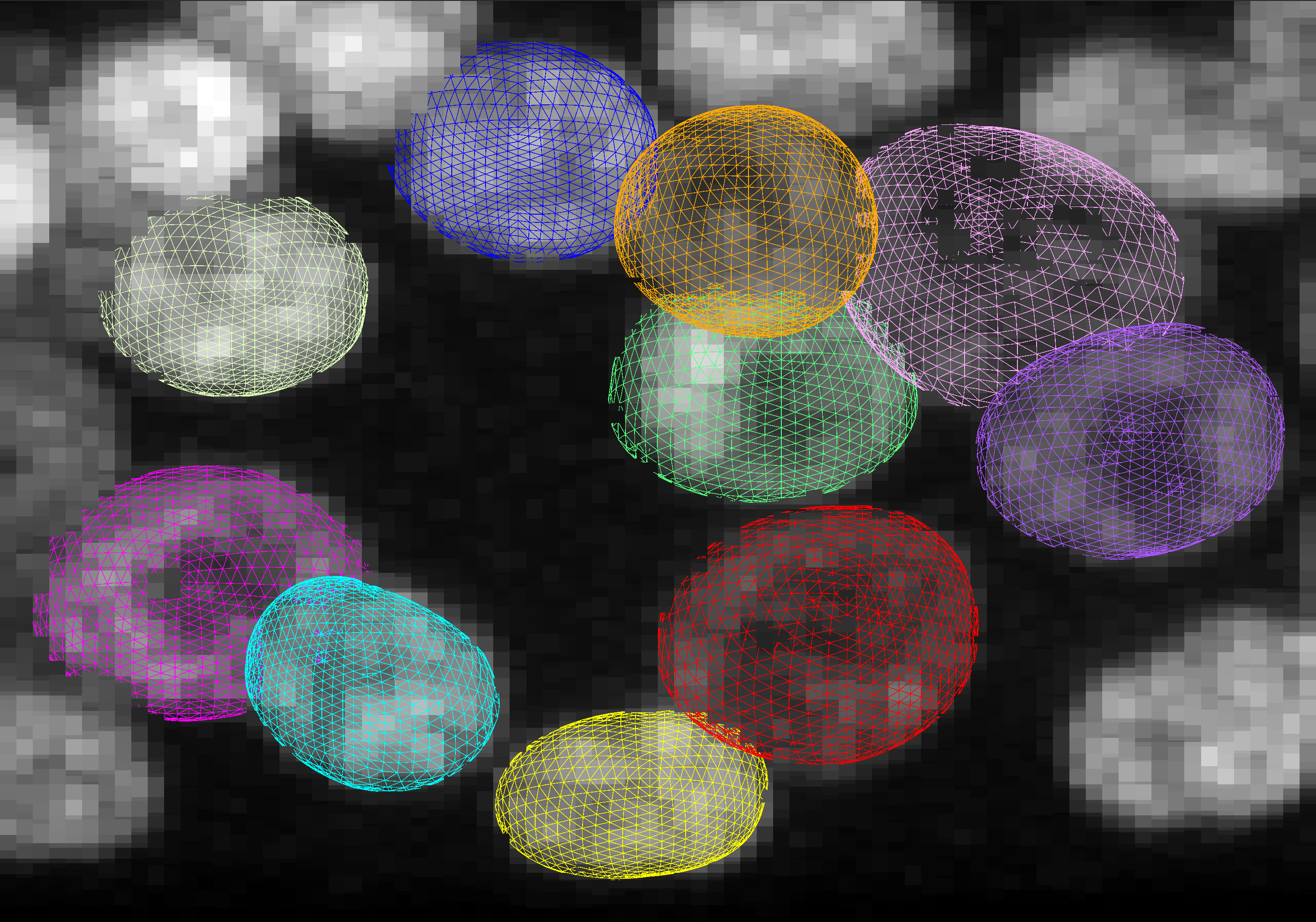}
        \label{fig:surf_wireframe1}
    \end{subfigure}
    \begin{subfigure}{0.475\textwidth}
        \includegraphics[width=\textwidth]{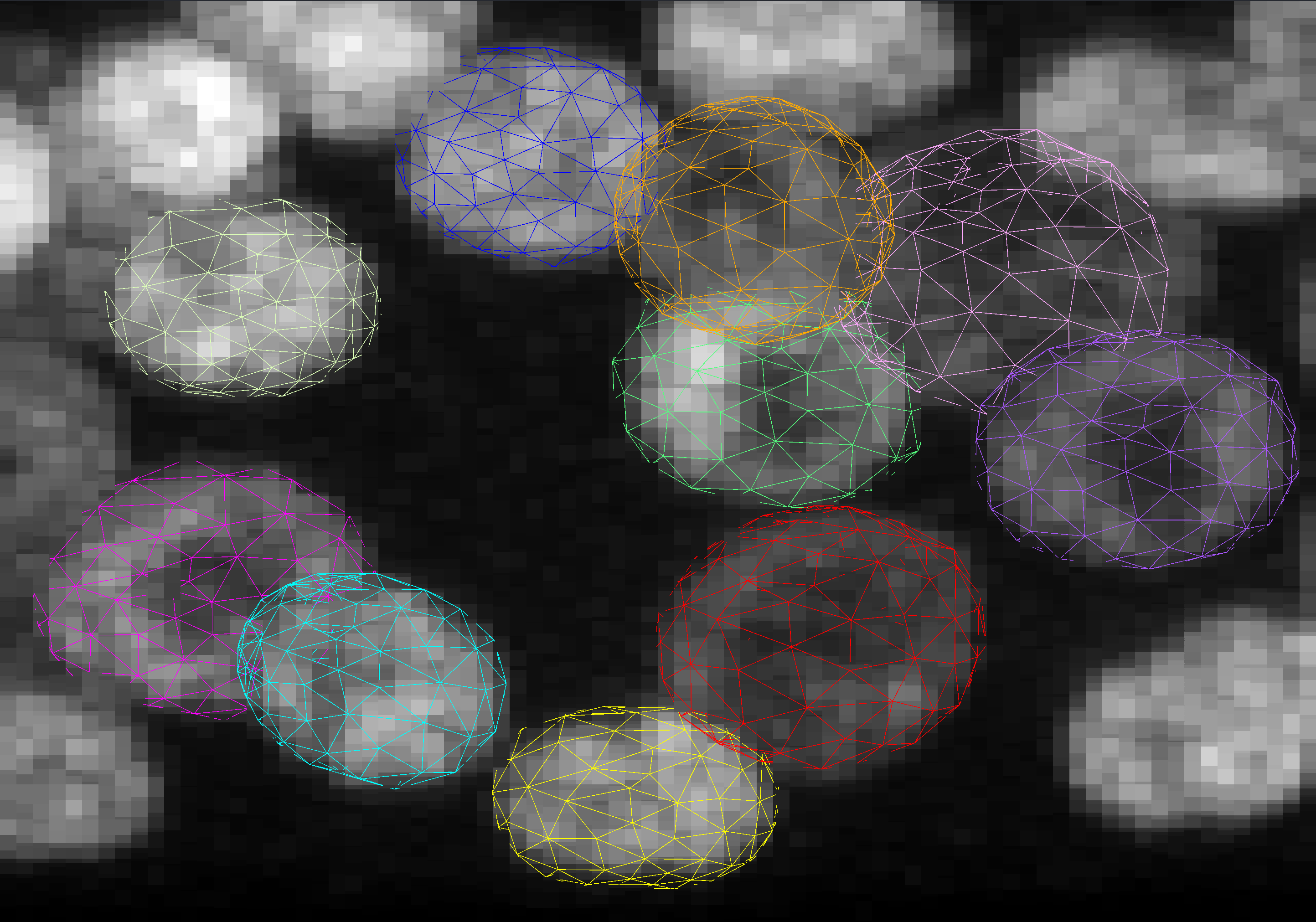}
        \label{fig:star_wireframe1}
    \end{subfigure}
    \caption{
    SurfDist's parameterization produces smoothly curved instances.
    Wireframes learned by 6-ray SurfDist (left) and 96-ray StarDist-3D (right) are rendered for arbitrarily selected instances in the dataset \textit{Worm} from an arbitrary viewing angle. Contrast-adjusted input images are rendered in grayscale. Predicted wireframes are colored randomly. Renderings were generated using \cite{https://doi.org/10.5281/zenodo.3555620}.}
    \label{fig:wireframes}
\end{figure*}

\textit{Worm} was chosen as a real-data benchmark, following \cite{weigert2020star}. We trained five StarDist-3D and ten SurfDist models with an 18/3 train/validation split. To match the recommended default parameterization for StarDist-3D and the parameterization for which results on \textit{Worm} were reported in \cite{weigert2020star}, we used a 96-ray StarDist-3D model for all five StarDist-3D training runs. For SurfDist, to demonstrate its relative compactness, we trained five models each for two parameterizations: one using 6 rays (38 total free parameters) per instance and one using 12 rays (92 total free parameters) per instance. Comparison of validation segmentation accuracy of trained models (\cref{table:worm_metrics}) shows that 6-ray SurfDist models performed comparably to 96-ray StarDist-3D models on average, and our best 6-ray SurfDist model outperformed all 96-ray StarDist-3D models. 12-ray SurfDist models outperformed 96-ray StarDist-3D models both on average and in the best case. Notably, instance resolution in this dataset is relatively low, a feature which should attenuate SurfDist's advantage over StarDist-3D. Example renderings of meshes predicted for this dataset by one of our trained SurfDist models are shown in \cref{fig:wireframes}.

\begin{table}
    \begin{center}
    \begin{tabular}{ lrrrr }
        \hline
        \multicolumn{5}{c}{Across Five Trained Models} \\
        \hline
        Model & Precision & Recall & Accuracy & F1\\
        \hline
        SurfDist 6 & .6495 & .6545 & .5683 & .6520\\
        SurfDist 12 & \textbf{.6510} & \textbf{.6617} & \textbf{.5722} & \textbf{.6563}\\
        StarDist-3D 96 & .6141 & .6069 & .5285 & .6105\\
        \hline
        \multicolumn{5}{c}{Best Trained Model} \\
        \hline
        Model & Precision & Recall & Accuracy & F1\\
        \hline
        SurfDist 6 & \textbf{.6606} & \textbf{.6651} & \textbf{.5770} & \textbf{.6629}\\
        SurfDist 12 & .6489 & .6627 & .5719 & .6557\\
        StarDist-3D 96 & .6254 & .6117 & .5365 & .6185\\
        \hline
    \end{tabular}
    \end{center}
    \caption{
        Mean validation segmentation metrics for dataset \textit{Worm} of trained SurfDist and StarDist-3D models. For each model architecture, five models were trained for 2000 epochs on 96x64x64 patches with downsampling by a factor of 2. 18 volumes were used for training and 3 volumes for validation. Results displayed are means across IoU thresholds $(0.1, 0.2, 0.3, 0.4, 0.5, 0.6, 0.7, 0.8, 0.9)$. Numerical suffixes in model names indicate the number of rays used in model instances.
    }
    \label{table:worm_metrics}
\end{table}

\begin{figure*}
\centering
\begin{subfigure}{0.45\textwidth}
    \includegraphics[width=\textwidth]{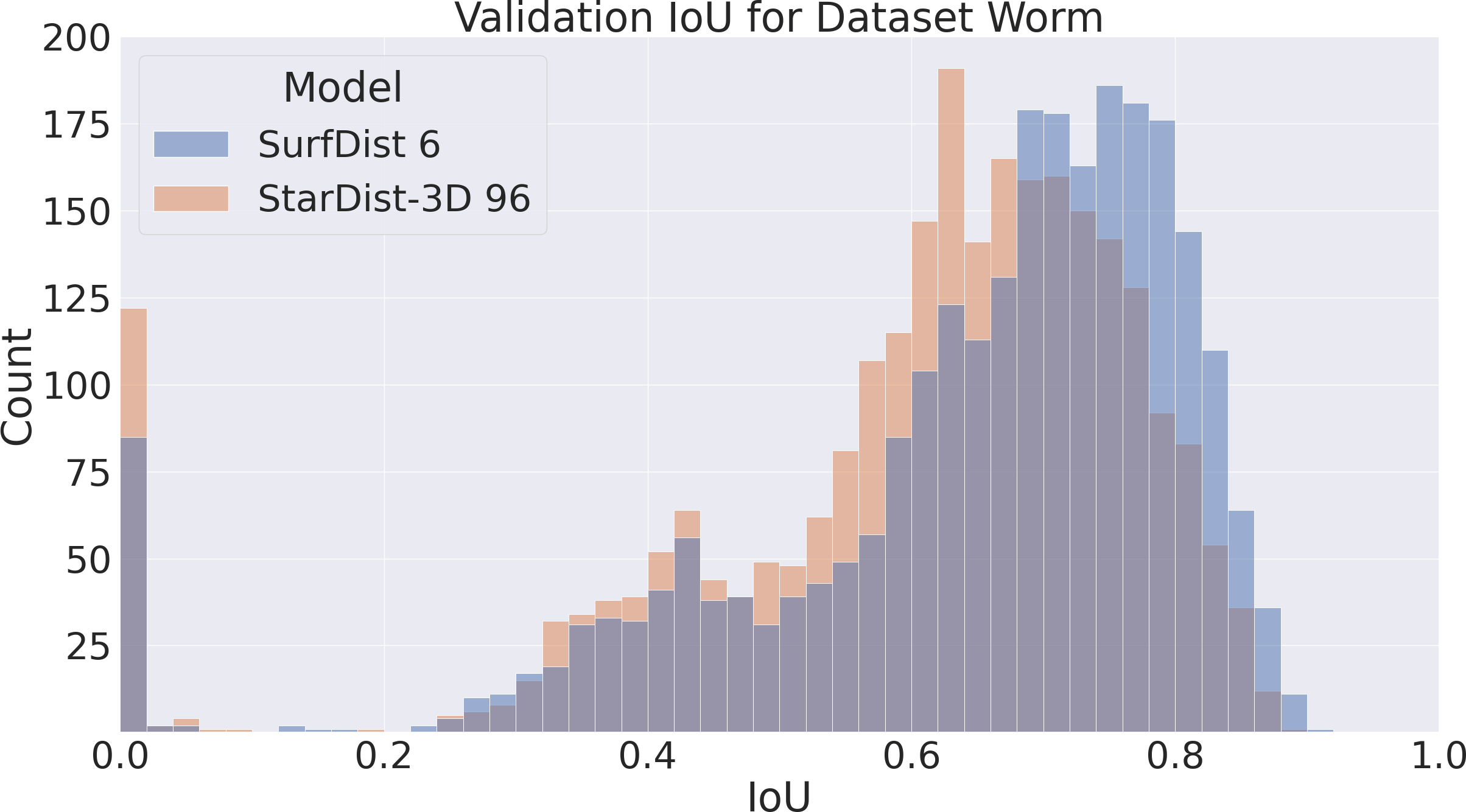}
    \label{fig:worm_ious_6}
\end{subfigure}
\begin{subfigure}{0.45\textwidth}
    \includegraphics[width=\textwidth]{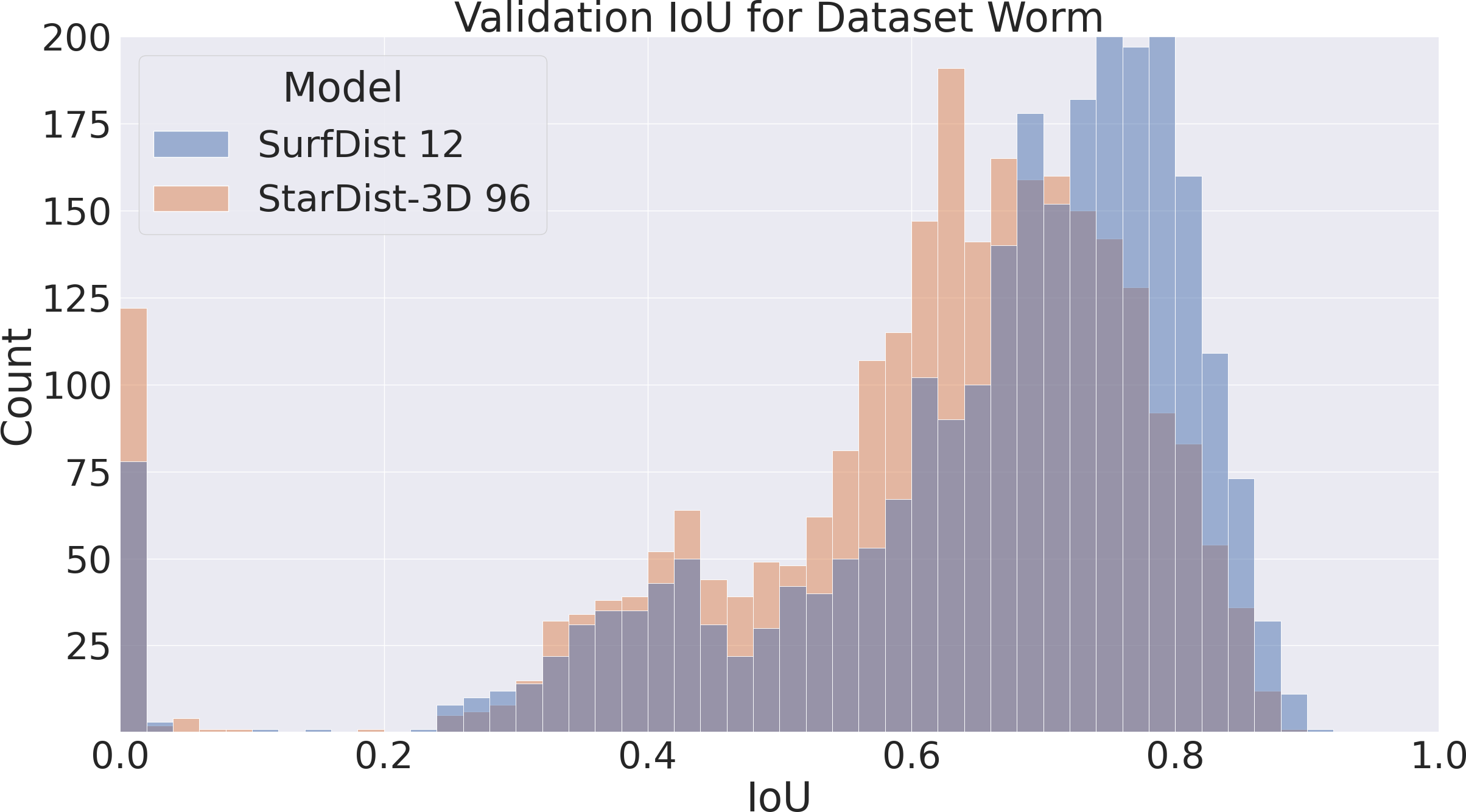}
    \label{fig:worm_ious_12}
\end{subfigure} 
\caption{
  IoU distributions for matched predicted and true validation instances for dataset \textit{Worm} across all trained models detailed in \cref{table:worm_metrics}. Models of type SurfDist 6 use 6 rays, 8 triangular faces, and 38 total free parameters per instance. Models of type SurfDist 12 use 12 rays, 20 triangular faces, and 92 total free parameters per instance. Models of type StarDist-3D 96 use 96 rays, 188 triangular faces, and 96 total free parameters per instance.
  }
\label{fig:worm_performance}
\end{figure*}

\subsection{Inference-Time Interpolation vs. Direct Learning of Parametric Surface Representations}

SurfDist models instances by directly learning parametric surface representations. StarDist-3D directly learns only sparse surface samplings, but these sparse surface samplings can be used to infer surface representations such as those learned by SurfDist. To test whether SurfDist actually learns different information than StarDist-3D, we modified StarDist-3D models trained on the \textit{Worm} dataset to use SurfDist's curved mesh construction at inference time. To define these curved surfaces, we calculated a normal direction for each mesh vertex as the average of its triangular faces, and we used these vertex normals alongside the vertex coordinates to infer a bicubic Bézier triangle for each flat mesh triangle with the point-normal triangle algorithm \cite{kato2002curved}.

As an implementation detail, SurfDist enforces equidistant ray spacing for 6-, 8-, and 12- ray models such that the mesh computed for the unit rays will form an octahedron, a cube (with two co-planar triangular faces per square face), or an icosahedron respectively. Equidistant spacings are enforced for these ray counts because we found that the Fibonacci lattice \cite{gonzalez2010measurement} approach produces highly irregular ray distributions for low ray counts. To disentangle the effects of our corrected ray distribution from those of our mesh construction, we trained StarDist-3D models using both the Fibonacci lattice approach which is typical for StarDist-3D and with the same equidistant ray directions as corresponding SurfDist models.

We find that while both the equidistant ray and point-normal triangle modifications improve the segmentation performance of 6- and 12-ray StarDist-3D models, 6- and 12-ray SurfDist models further outperform the modified StarDist-3D models \cref{table:ablation}. Interestingly, for the 96-ray StarDist-3D model, the point-normal triangle modification degrades rather than improves segmentation performance. This result may be explained by the intuition that when the number of rays in a StarDist-3D model is high enough for the meshes it produces to model local instance curvature well, inducing additional mesh complexity is likely to introduce error.

\begin{table}
    \begin{center}
    \begin{tabular}{ lcccc }
        \hline
        Model & Precision & Recall & Accuracy & F1\\
        \hline
        Star 6 & .2048 & .2030 & .1651 & .2039\\
        Star 6 EQ & .3571 & .3609 & .3069 & .3590\\
        Star 6 PN & .3887 & .3897 & .3269 & .3892\\
        Star 6 EQ PN & .5607 & .5663 & .4913 & .5635\\
        \hline
        Star 12 & .5551 & .5485 & .4793 & .5518\\
        Star 12 EQ & .5921 & .5947 & .5161 & .5934\\
        Star 12 PN & .6160 & .6202 & .5377 & .6181\\
        Star 12 EQ PN & .6169 & .6254 & .5407 & .6211\\
        \hline
        Star 96 & .6254 & .6117 & .5365 & .6185\\
        Star 96 PN & .6033 & .5935 & .5192 & .5983\\
        \hline
        SurfDist 6 & \textbf{.6606} & \textbf{.6651} & \textbf{.5770} & \textbf{.6629}\\
        SurfDist 12 & .6489 & .6627 & .5719 & .6557\\
        \hline
    \end{tabular}
    \end{center}
    \caption{Mean validation segmentation metrics for dataset \textit{Worm} of trained SurfDist models are compared to those of modified and unmodified trained StarDist3D models. "StarDist-3D" is abbreviated as "Star" to save table space. Numerical suffixes in model names indicate the number of rays used in model instances. "EQ" suffixes in StarDist-3D model names denote forced equidistant ray distributions (matching those used in SurfDist models for the same ray counts). "PN" suffixes in StarDist-3D model names denote use of a point-normal interpolation scheme (with a number of subdivisions at inference time matching the corresponding SurfDist model) to replace flat triangular mesh facets with bicubic Bézier triangles. Model thresholding was performed separately for each StarDist-3D variant prior to computation of segmentation metrics.}
    \label{table:ablation}
\end{table}

\subsection{Runtime Analysis}

\cref{table:runtime} compares training and inference run times observed for SurfDist and StarDist-3D during our experiments on the \textit{Worm} dataset. Generally, SurfDist models required ~10x as much time for training as StarDist-3D models. This slowdown is a consequence of SurfDist's more complicated loss function implementation, which could likely be improved through TensorFlow code optimization.

For inference, our best SurfDist models are ~50\% faster than a 96-ray StarDist-3D model. We note that the number of mesh subdivisions completed during inference is a specifiable hyperparameter. Inference time for SurfDist models scales supralinearly in the number of subdivisions, as each subdivision results in three new vertices per mesh triangle and results in a multiplication of the number of triangles in the mesh by a factor of four, and each vertex in the fully subdivided mesh is ultimately passed to StarDist-3D's inference routine as an individual ray. We found, unexpectedly, that using only 1 or 2 inference subdivisions produced better segmentation results for the \textit{Worm} dataset than using 3 or 4 \cref{table:inference_subdivisions}. This phenomenon may be a consequence of the low resolution of this dataset or of the widespread presence of edge artifacts in its annotated ground truth labels. Regardless, using 1 or 2 subdivisions results in an inference time advantage for SurfDist relative to StarDist-3D on this dataset.

\begin{table}
    \begin{center}
    \begin{tabular}{ lcc }
        \hline
        Model & epoch (Training) & image (Inference) \\
        \hline
        StarDist-3D 6 & \textbf{2.16} & \textbf{8.02}\\
        StarDist-3D 12 & 2.29 & 8.56\\
        StarDist-3D 96 & 3.35 & 29.61\\
        SurfDist 6 & 11.31 & 21.06\\
        SurfDist 12 & 27.26 & 20.77\\
        \hline
    \end{tabular}
    \end{center}
    \caption{Mean observed run times in seconds for training and inference on dataset \textit{Worm}. Patch sizing of $(92,  64, 64)$ was used for all models. For SurfDist models, two subdivisions were used during training and two (6-ray model) or one (12-ray model) subdivisions were used during inference.}
    \label{table:runtime}
\end{table}

\begin{table}
    \begin{center}
    \begin{tabular}{ lccc }
        \hline
        Model & Subdivisions & image (Inference) & F1 \\
        \hline
        SurfDist 6 & 4 & 166.05 & .6331\\
        SurfDist 6 & 3 & 59.17 & .6522\\
        SurfDist 6 & 2 & 21.06 & \textbf{.6629}\\
        SurfDist 6 & 1 & \textbf{10.91} & .6071\\
        SurfDist 12 & 4 & 481.61 & .6285\\
        SurfDist 12 & 3 & 105.42 & .6358\\
        SurfDist 12 & 2 & 43.79 & .6424\\
        SurfDist 12 & 1 & 20.77 & .6557\\
        \hline
    \end{tabular}
    \end{center}
    \caption{Mean observed runtimes in seconds for inference as number of inference subdivisions varies on dataset \textit{Worm}. Patch sizing of $(92,  64, 64)$ was used for all models.}
    \label{table:inference_subdivisions}
\end{table}

\subsection{Segmentation of NucMM and Parhyale Volumetric Microscopy Datasets}

To test SurfDist's ability to generalize beyond the well-rounded nuclei in the \textit{Worm} dataset, we trained models for two other benchmarks: \textit{NucMM} \cite{lin2021nucmm}, an electron microscopy volume of a zebrafish brain, and \textit{Parhyale} \cite{alwes_2023_8252039}, a volumetric confocal microscopy timeseries of regenerating \textit{Parhyale hawaiensis} legs. These datasets were previously used as benchmarks in \cite{weigert2020star} (\textit{Parhyale}) and \cite{wu2023nisnet3d} (\textit{NucMM}).

\begin{table}
    \begin{center}
    \begin{tabular}{ lcccc }
        \hline
        Model & Precision & Recall & Accuracy & F1\\
        \hline
        StarDist-3D 96 & .9773 & .5639 & .5566 & .7151\\
        SurfDist 6 & .8313 & .4775 & .4482 & .6066\\
        SurfDist 8 & \textbf{.9849} & .5627 & .5579 & .7162\\
        SurfDist 12 & .8427 & .4832 & .4555 & .6143\\
        \hline
        Cellpose & .9615 & .9447 & \text{N/A} & .9530\\
        NisNet3D & .9689 & \textbf{.9624} & \text{N/A} & \textbf{.9656}\\
        \hline
    \end{tabular}
    \end{center}
    \caption{Mean validation segmentation metrics across IoU thresholds $(0.5, 0.55, 0.6, 0.65, 0.7, 0.75)$ after 100 epochs of training on dataset \textit{NucMM}. Metrics for Cellpose \cite{stringer2021cellpose} and NisNet3D \cite{wu2023nisnet3d} are reprinted here from \cite{wu2023nisnet3d} for comparison. All SurfDist models used 2 subdivisions at training time and 4 subdivisions at inference time. All models processed entire $(64,64,64)$ subvolumes without further chunking.}
    \label{table:nucmm}
\end{table}

\begin{figure}
\centering
\begin{subfigure}{0.475 \textwidth}
    \includegraphics[width=\textwidth]{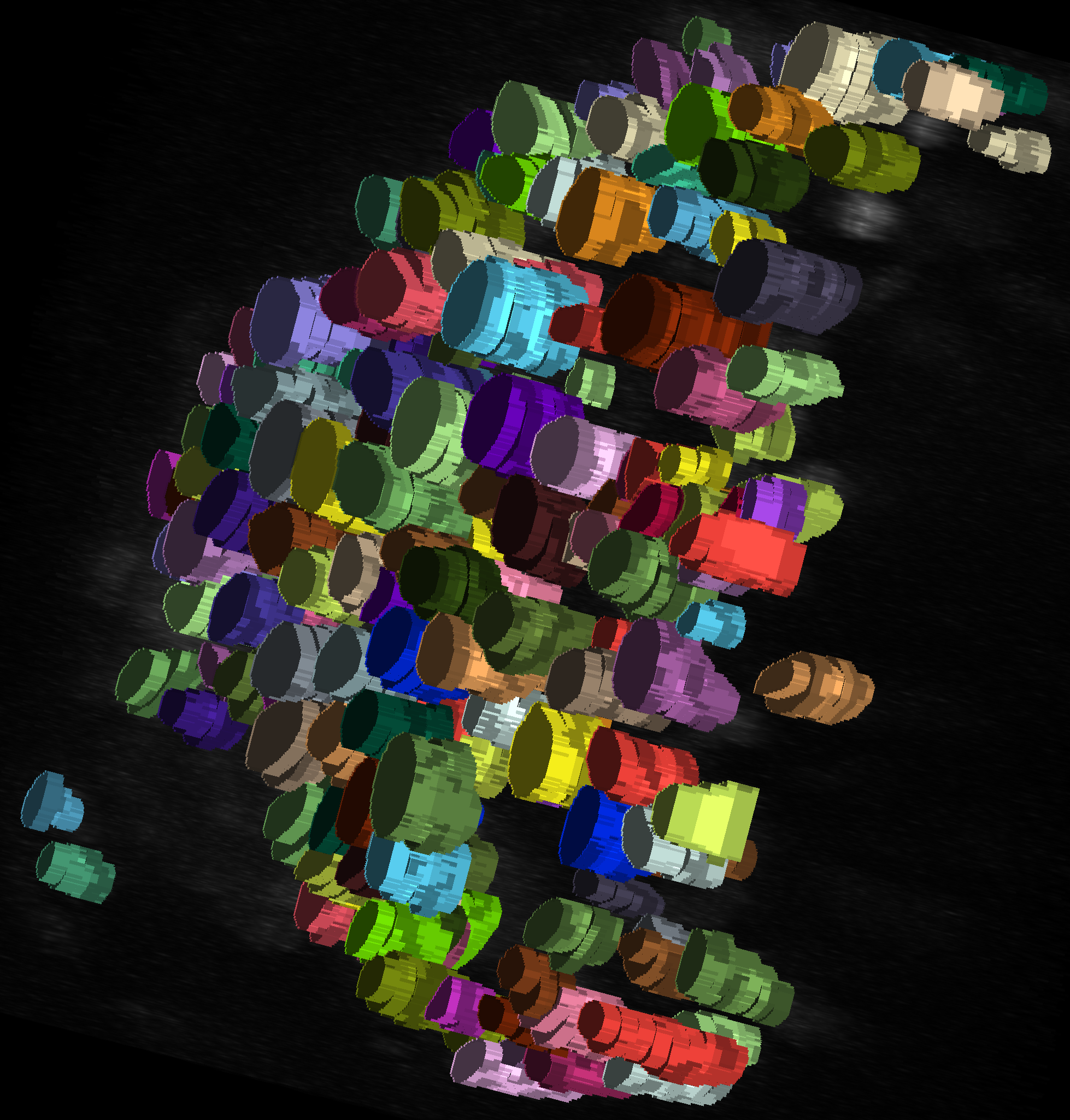}
    \caption{Grayscale image and pseudocolored ground truth labels for volume \textit{tp01} of dataset \textit{Parhyale}. Scaling factors of $(8, 1, 1)$ are used for both images.}
    \label{fig:parhyale_gt}
\end{subfigure}
\begin{subfigure}{0.475 \textwidth}
    \includegraphics[width=\textwidth]{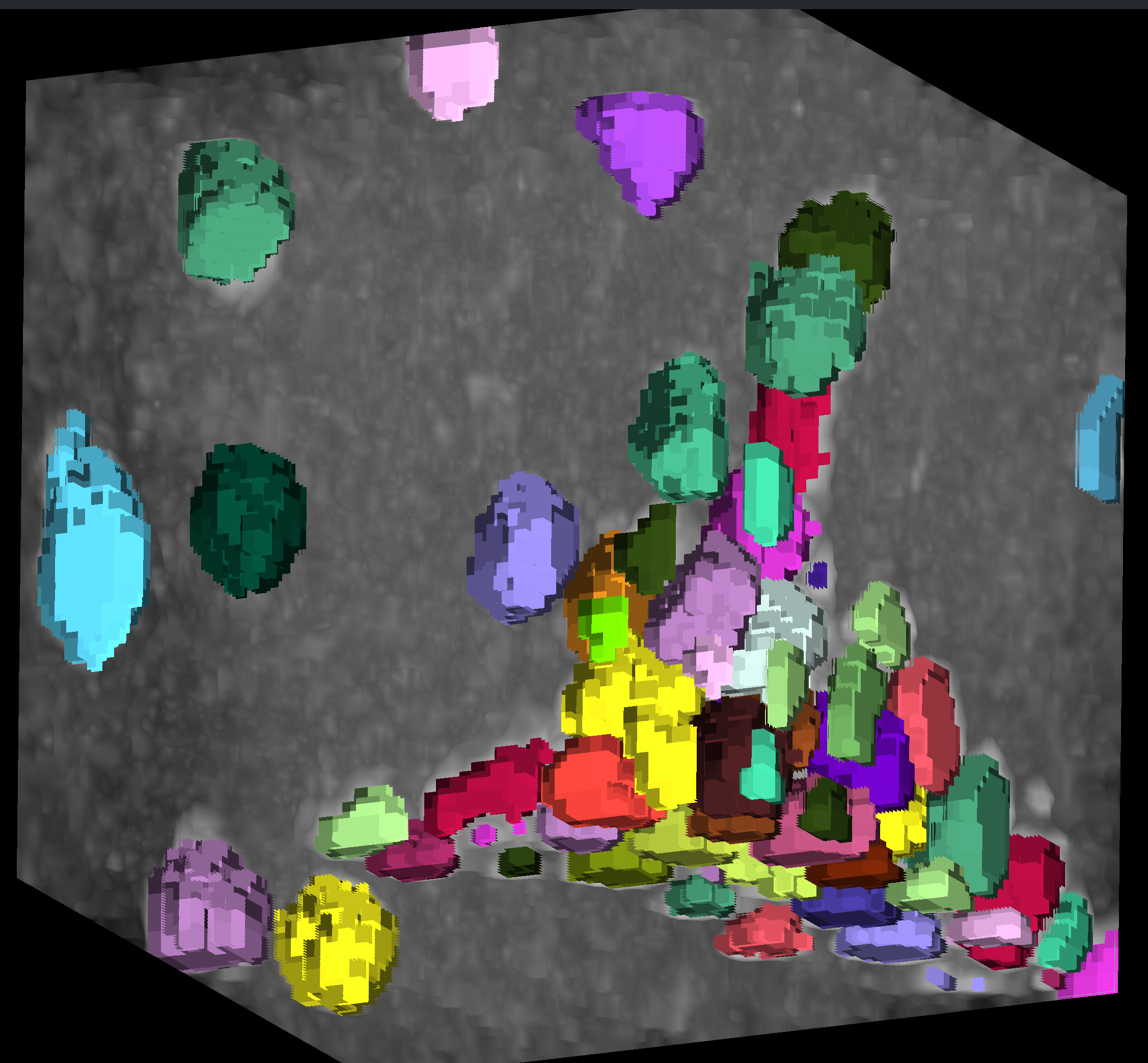}
    \caption{Grayscale image and pseudocolored ground truth labels for volume \textit{0000\_0576\_0768} of dataset \textit{NucMM}.}
    \label{fig:nucmm_gt}
\end{subfigure}
\caption{Representative volumes of datasets \textit{Parhyale} and \textit{NucMM}. Each volume shown here is the first training volume of its dataset. Renderings were generated using \cite{https://doi.org/10.5281/zenodo.3555620}.}
\label{fig:non-biological_artifiacts}
\end{figure}

After training 6- and 12-ray SurfDist models on \textit{NucMM}, we observed poor segmentation performance relative both to StarDist-3D models which we trained and to the results reported in \cite{wu2023nisnet3d} for the alternative 3D segmentation approaches NISNet3D \cite{wu2023nisnet3d} and Cellpose \cite{stringer2021cellpose} (\cref{table:nucmm}). We hypothesized that this was a consequence of the clipping of a high proportion of training instances at subvolume boundaries, resulting from of the small size ($64\text{x}64\text{x}64\text{ voxels}$) of each annotated subvolume in the dataset (\cref{fig:nucmm_gt}). Clipping artifacts are likely to be modeled better by StarDist-3D, NISNet3D, and Cellpose, none of which explicitly assume curved instance surfaces, than by SurfDist. To validate this hypothesis, we defined an equidistant distribution for 8-ray SurfDist models such that the mesh of the unit rays forms an axes-aligned cube with two co-planar triangles per square face. We reasoned that models trained using this ray definition should handle subvolume clipping artifacts more gracefully than the non-axes aligned octahedral or icosahedral meshes of our 6- or 12-ray models. Results observed after training an 8-ray model validated this: our 8-ray SurfDist model outperformed not only our 6- and 12-ray SurfDist models, but also the 96-ray StarDist-3D model that we trained for comparison. Our 8-ray SurfDist model is still outperformed by NISNet3D and Cellpose on this dataset, but this is neither a surprising nor a damning result, because NISNet3D and Cellpose have greater freedom to optimize segmentation results, since they are not restricted like SurfDist or StarDist-3D by a requirement to learn parametric instance representations. Additionally, NISNet3D and Cellpose are built on more sophisticated neural network backbones than the standard 3D U-Net \cite{cicek20163dunetlearningdense} used in our SurfDist and StarDist-3D models. These backbones should also be compatible with SurfDist and StarDist-3D, suggesting an area of future work.

\begin{table}
    \begin{center}
    \begin{tabular}{ lcccc }
        \hline
        Model & Precision & Recall & Accuracy & F1\\
        \hline
        SurfDist 6 & .5152 & .4795 & .3976 & .4967\\
        SurfDist 8 & .5416 & .5153 & .4252 & .5281\\
        SurfDist 12 & .5388 & \textbf{.5247} & .4316 & .5317\\
        StarDist-3D 96 & \textbf{.5757} & .5128 & \textbf{.4436} & .\textbf{5424}\\
        \hline
    \end{tabular}
    \end{center}
    \caption{Mean validation segmentation metrics across IoU thresholds $(0.1, 0.2, 0.3, 0.4, 0.5, 0.6, 0.7, 0.8, 0.9)$ after 2000 epochs of training on dataset \textit{Parhyale}. All SurfDist models used 2 subdivisions at training time and 4 subdivisions at inference time. All models processed patches of size $(32,64,64)$.}
    \label{table:parhyale}
\end{table}

For \textit{Parhyale}, we observed similar poorer SurfDist performance relative to StarDist-3D. We hypothesize a similar explanation: human labeling artifacts in the training dataset result in ground truth annotations which appear unnaturally cylindrical and do not accurately label voxels near the boundaries of cell nuclei (\cref{fig:parhyale_gt}). Despite the ill-suitedness of these cylindrical-like label shapes to SurfDist's mesh construction, our trained SurfDist models produce segmentations which are comparable to those produced by our trained StarDist-3D models (\cref{table:parhyale}). We note that where StarDist-3D outperforms SurfDist for this dataset, it does so only marginally, and likely by reproducing edge artifacts which do not accurately reflect the underlying data. In such cases, SurfDist's lower-scoring segmentations, which induce curved instance boundaries, may actually provide more realistic instance boundaries.

\section{Discussion}

SurfDist learns representations of three-dimensional instances as closed meshes of parameterized curved surface patches. Our experimental results demonstrate that it can achieve accuracy superior to that of StarDist-3D with a smaller number of parameters for blob-shaped objects which are common in biomedical imaging data. As the resolution of microscopy increases through technological development, we expect the relative advantage of curved surface modeling over faceted surface models to increase. 

Our tests show that the StarDist architectural approach can be effective applied to estimate parametric surfaces underlying 3D blob-like data, and it motivates further incorporation of computational geometry approaches for three-dimensional instance segmentation. For example, SurfDist could be extended to support more complex surface topologies.

Despite its state-of-the-art performance on segmentation tasks, SurfDist has two significant limitations which should inform the direction of future work. First, like StarDist-3D, SurfDist is ill-suited for datasets where instances are not star-convex, such as branching dendritic cells. Second, additional work is needed to reduce the computational time required for inference using trained SurfDist (or StarDist-3D) models. The current bottleneck in inference is the non-maximum suppression step, in which an instance is predicted for every voxel of the input volume and pairwise intersections of predicted instances are computed to de-duplicate predictions. Computation of intersections of objects bounded by polyhedral meshes is a common task in computer graphics and already highly optimized, so a fruitful avenue for reducing inference costs may be development of alternative schemes for non-maximum suppression that decouple instance predictions from input voxel grids.

Finally, our exploration of existing commonly used reference 3D datasets revealed obvious and substantial artifacts in their ground-truth annotations. Limited availability of high-quality training data has long been a serious limiting factor on the advancement of 3D image segmentation approaches, and though efforts toward publication of new datasets have increased encouragingly in recent years, our results indicate that work on dataset generation and curation is still sorely needed. We hope that our results will help to motivate continued work in these areas.
\section{Data and code availability}

Code for the SurfDist model and for reproduction of the experiments detailed here is available on GitHub.
\clearpage
{
    \small
    \bibliographystyle{ieeenat_fullname}
    \bibliography{main}

\begin{thebibliography}{16}
\providecommand{\natexlab}[1]{#1}
\providecommand{\url}[1]{\texttt{#1}}
\expandafter\ifx\csname urlstyle\endcsname\relax
  \providecommand{\doi}[1]{doi: #1}\else
  \providecommand{\doi}{doi: \begingroup \urlstyle{rm}\Url}\fi

\bibitem[Alwes et~al.(2023)Alwes, Sugawara, and Averof]{alwes_2023_8252039}
Frederike Alwes, Ko Sugawara, and Michalis Averof.
\newblock Parhyale 3d segmentation dataset, 2023.

\bibitem[Gonz{\'a}lez(2010)]{gonzalez2010measurement}
{\'A}lvaro Gonz{\'a}lez.
\newblock Measurement of areas on a sphere using fibonacci and latitude--longitude lattices.
\newblock \emph{Mathematical geosciences}, 42:\penalty0 49--64, 2010.

\bibitem[Kato(2002)]{kato2002curved}
Saul Kato.
\newblock Curved surface reconstruction, 2002.
\newblock US Patent 6,462,738.

\bibitem[Lee et~al.(2016)Lee, Hwang, and Yoon]{lee2016bezier}
Chang-Ki Lee, Hae-Do Hwang, and Seung-Hyun Yoon.
\newblock B{\'e}zier triangles with g 2 continuity across boundaries.
\newblock \emph{Symmetry}, 8\penalty0 (3):\penalty0 13, 2016.

\bibitem[Lin et~al.(2021)Lin, Wei, Petkova, Wu, Ahmed, K, Zou, Wendt, Boulanger-Weill, Wang, et~al.]{lin2021nucmm}
Zudi Lin, Donglai Wei, Mariela~D Petkova, Yuelong Wu, Zergham Ahmed, Krishna~Swaroop K, Silin Zou, Nils Wendt, Jonathan Boulanger-Weill, Xueying Wang, et~al.
\newblock Nucmm dataset: 3d neuronal nuclei instance segmentation at sub-cubic millimeter scale.
\newblock In \emph{International Conference on Medical Image Computing and Computer-Assisted Intervention}, pages 164--174. Springer, 2021.

\bibitem[Long et~al.(2009)Long, Peng, Liu, Kim, and Myers]{long20093d}
Fuhui Long, Hanchuan Peng, Xiao Liu, Stuart~K Kim, and Eugene Myers.
\newblock A 3d digital atlas of c. elegans and its application to single-cell analyses.
\newblock \emph{Nature methods}, 6\penalty0 (9):\penalty0 667--672, 2009.

\bibitem[Mandal and Uhlmann(2021)]{mandal2021splinedist}
Soham Mandal and Virginie Uhlmann.
\newblock Splinedist: Automated cell segmentation with spline curves.
\newblock In \emph{2021 IEEE 18th International Symposium on Biomedical Imaging (ISBI)}, pages 1082--1086. IEEE, 2021.

\bibitem[Schmidt et~al.(2018)Schmidt, Weigert, Broaddus, and Myers]{schmidt2018cell}
Uwe Schmidt, Martin Weigert, Coleman Broaddus, and Gene Myers.
\newblock Cell detection with star-convex polygons.
\newblock In \emph{Medical image computing and computer assisted intervention--MICCAI 2018: 21st international conference, Granada, Spain, September 16-20, 2018, proceedings, part II 11}, pages 265--273. Springer, 2018.

\bibitem[Sofroniew et~al.(2025)Sofroniew, Lambert, Bokota, Nunez-Iglesias, Sobolewski, Sweet, Gaifas, Evans, Burt, Doncila~Pop, Yamauchi, Weber~Mendon\c{c}a, Liu, Buckley, Vierdag, Monko, Royer, Can~Solak, Harrington, Ahlers, Althviz~Moré, Amsalem, Anderson, Annex, Aronssohn, Boone, Bragantini, Bussonnier, Caporal, Eglinger, Eisenbarth, Freeman, Gohlke, Gunalan, Halchenko, Har-Gil, Harfouche, Hilsenstein, Hutchings, Lauer, Lichtner, Liu, Liu, Lowe, Marconato, Martin, McGovern, Migas, Miller, Miñano, Muñoz, M\"{u}ller, Nauroth-Kreß, Obenhaus, Palecek, Pape, Perlman, Pevey, Peña-Castellanos, Pierré, Pinto, Rodríguez-Guerra, Ross, Russell, Ryan, Selzer, Smith, Smith, Sofiiuk, Soltwedel, Stansby, Vanaret, Wadhwa, Weigert, Willing, Windhager, Winston, and Zhao]{https://doi.org/10.5281/zenodo.3555620}
Nicholas Sofroniew, Talley Lambert, Grzegorz Bokota, Juan Nunez-Iglesias, Peter Sobolewski, Andrew Sweet, Lorenzo Gaifas, Kira Evans, Alister Burt, Draga Doncila~Pop, Kevin Yamauchi, Melissa Weber~Mendon\c{c}a, Lucy Liu, Genevieve Buckley, Wouter-Michiel Vierdag, Timothy Monko, Loic Royer, Ahmet Can~Solak, Kyle I.~S. Harrington, Jannis Ahlers, Daniel Althviz~Moré, Oren Amsalem, Ashley Anderson, Andrew Annex, Constantin Aronssohn, Peter Boone, Jordão Bragantini, Matthias Bussonnier, Clément Caporal, Jan Eglinger, Andreas Eisenbarth, Jeremy Freeman, Christoph Gohlke, Kabilar Gunalan, Yaroslav~Olegovich Halchenko, Hagai Har-Gil, Mark Harfouche, Volker Hilsenstein, Katherine Hutchings, Jessy Lauer, Gregor Lichtner, Hanjin Liu, Ziyang Liu, Alan Lowe, Luca Marconato, Sean Martin, Abigail McGovern, Lukasz Migas, Nadalyn Miller, Sofía Miñano, Hector Muñoz, Jan-Hendrik M\"{u}ller, Christopher Nauroth-Kreß, Horst~A. Obenhaus, David Palecek, Constantin Pape, Eric Perlman, Kim Pevey, Gonzalo Peña-Castellanos,
  Andrea Pierré, David Pinto, Jaime Rodríguez-Guerra, David Ross, Craig~T. Russell, James Ryan, Gabriel Selzer, MB Smith, Paul Smith, Konstantin Sofiiuk, Johannes Soltwedel, David Stansby, Jules Vanaret, Pam Wadhwa, Martin Weigert, Carol Willing, Jonas Windhager, Philip Winston, and Rubin Zhao.
\newblock napari: a multi-dimensional image viewer for python, 2025.

\bibitem[Stringer et~al.(2021)Stringer, Wang, Michaelos, and Pachitariu]{stringer2021cellpose}
Carsen Stringer, Tim Wang, Michalis Michaelos, and Marius Pachitariu.
\newblock Cellpose: a generalist algorithm for cellular segmentation.
\newblock \emph{Nature methods}, 18\penalty0 (1):\penalty0 100--106, 2021.

\bibitem[van~der Walt et~al.(2014)van~der Walt, {S}ch\"onberger, {Nunez-Iglesias}, {B}oulogne, {W}arner, {Y}ager, {G}ouillart, {Y}u, and the scikit-image contributors]{scikit-image}
{S}t\'efan van~der Walt, {J}ohannes~{L}. {S}ch\"onberger, {J}uan {Nunez-Iglesias}, {F}ran\c{c}ois {B}oulogne, {J}oshua~{D}. {W}arner, {N}eil {Y}ager, {E}mmanuelle {G}ouillart, {T}ony {Y}u, and the scikit-image contributors.
\newblock scikit-image: image processing in {P}ython.
\newblock \emph{PeerJ}, 2:\penalty0 e453, 2014.

\bibitem[Virtanen et~al.(2020)Virtanen, Gommers, Oliphant, Haberland, Reddy, Cournapeau, Burovski, Peterson, Weckesser, Bright, {van der Walt}, Brett, Wilson, Millman, Mayorov, Nelson, Jones, Kern, Larson, Carey, Polat, Feng, Moore, {VanderPlas}, Laxalde, Perktold, Cimrman, Henriksen, Quintero, Harris, Archibald, Ribeiro, Pedregosa, {van Mulbregt}, and {SciPy 1.0 Contributors}]{2020SciPy-NMeth}
Pauli Virtanen, Ralf Gommers, Travis~E. Oliphant, Matt Haberland, Tyler Reddy, David Cournapeau, Evgeni Burovski, Pearu Peterson, Warren Weckesser, Jonathan Bright, St{\'e}fan~J. {van der Walt}, Matthew Brett, Joshua Wilson, K.~Jarrod Millman, Nikolay Mayorov, Andrew R.~J. Nelson, Eric Jones, Robert Kern, Eric Larson, C~J Carey, {\.I}lhan Polat, Yu Feng, Eric~W. Moore, Jake {VanderPlas}, Denis Laxalde, Josef Perktold, Robert Cimrman, Ian Henriksen, E.~A. Quintero, Charles~R. Harris, Anne~M. Archibald, Ant{\^o}nio~H. Ribeiro, Fabian Pedregosa, Paul {van Mulbregt}, and {SciPy 1.0 Contributors}.
\newblock {{SciPy} 1.0: Fundamental Algorithms for Scientific Computing in Python}.
\newblock \emph{Nature Methods}, 17:\penalty0 261--272, 2020.

\bibitem[Weigert et~al.(2020)Weigert, Schmidt, Haase, Sugawara, and Myers]{weigert2020star}
Martin Weigert, Uwe Schmidt, Robert Haase, Ko Sugawara, and Gene Myers.
\newblock Star-convex polyhedra for 3d object detection and segmentation in microscopy.
\newblock In \emph{Proceedings of the IEEE/CVF winter conference on applications of computer vision}, pages 3666--3673, 2020.

\bibitem[Wu et~al.(2023)Wu, Chen, Salama, Winfree, Dunn, and Delp]{wu2023nisnet3d}
Liming Wu, Alain Chen, Paul Salama, Seth Winfree, Kenneth~W Dunn, and Edward~J Delp.
\newblock Nisnet3d: Three-dimensional nuclear synthesis and instance segmentation for fluorescence microscopy images.
\newblock \emph{Scientific Reports}, 13\penalty0 (1):\penalty0 9533, 2023.

\bibitem[Yamaguchi(1988)]{alma991085856964006532}
Fujio Yamaguchi.
\newblock \emph{Curves and surfaces in computer aided geometric design}.
\newblock Springer-Verlag, Berlin, 1st ed. 1988. edition, 1988.

\bibitem[Özgün Çiçek et~al.(2016)Özgün Çiçek, Abdulkadir, Lienkamp, Brox, and Ronneberger]{cicek20163dunetlearningdense}
Özgün Çiçek, Ahmed Abdulkadir, Soeren~S. Lienkamp, Thomas Brox, and Olaf Ronneberger.
\newblock 3d u-net: Learning dense volumetric segmentation from sparse annotation, 2016.

\end{thebibliography}
}

\end{document}